\documentclass[wcp]{jmlr}
\usepackage{amssymb}
\usepackage{amsmath}
\usepackage[ruled]{algorithm2e}
\usepackage{float}

\jmlrvolume{230}
\jmlryear{2024}
\jmlrworkshop{Conformal and Probabilistic Prediction with Applications}
\jmlrproceedings{PMLR}{Proceedings of Machine Learning Research}

\title{Adaptive Conformal Inference for Multi-Step Ahead Time-Series Forecasting Online}
\author{\Name{Johan Hallberg Szabadv\'ary}\Email{johan.hallberg.szabadvary@ju.se}\\
\addr{Department of Computing, J\"onk\"oping University, Sweden}}

\editor{Simone Vantini, Matteo Fontana, Aldo Solari, Henrik Boström and Lars Carlsson}

\begin{document}

\maketitle

\begin{abstract}
The aim of this paper is to propose an adaptation of the well known adaptive conformal inference (ACI) algorithm to achieve finite-sample coverage guarantees in multi-step ahead time-series forecasting in the online setting. ACI dynamically adjusts significance levels, and comes with finite-sample guarantees on coverage, even for non-exchangeable data. Our multi-step ahead ACI procedure inherits these guarantees at each prediction step, as well as for the overall error rate. The multi-step ahead ACI algorithm can be used with different target error and learning rates at different prediction steps, which is illustrated in our numerical examples, where we employ a version of the confromalised ridge regression algorithm, adapted to multi-input multi-output forecasting. The examples serve to show how the method works in practice, illustrating the effect of variable target error and learning rates for different prediction steps, which suggests that a balance may be struck between efficiency (interval width) and coverage.
\end{abstract}


\section{Introduction}

Conformal prediction (CP), described in detail in \cite{shafer2008tutorial} and \cite{alrw2}, is a general method for distribution-free uncertainty quantification, using past experience to determine precise levels of confidence in a new prediction. It can be applied to essentially any machine learning method, and has guaranteed properties of validity, provided that the data is drawn from a probability distribution that is exchangeable (see \cite{alrw2}, Section 2.1.1 for details), meaning essentially that any permutation is equally probable. In the online setting, where Reality presents us with successive examples $z_t:=(x_t, y_t)\in Z:=X\times Y$, each one consisting of an object $x_t\in X$ and its associated label $y_t\in Y$, CP produces prediction sets, $\Gamma_t^{\varepsilon}$ (typically intervals), at a user specified significance level $\varepsilon$, using $z_1,\dots, z_{t-1}$ and $x_t$ as input. They are valid in the sense that the true label, $y_t$, is contained in the prediction set with probability at least $1-\varepsilon$, and the error events are independent.\footnote{“at least” can be replaced by “exactly” by using smoothed CP, which we do not consider here.}
While powerful and highly general, the validity guarantees are lost if the exchangeability assumption is violated. This means in particular that CP comes with no guarantees for most time-series forecasting problems.

Adaptive conformal inference (ACI), introduced in \cite{gibbs2021adaptive}, is a method designed to achieve a desired coverage frequency $1-\varepsilon$ even in cases when the data is not exchangeable. Because ACI does not require data to be exchangeable, it has been suggested as a good candidate for producing prediction intervals in time-series forecasting \citep{zaffran2022adaptive}. It achieves, at least asymptotically, a user specified target error rate $\varepsilon$, by using a simple online update of the significance level
\begin{equation*}
    \label{eq:aciUpdateSingle}
    \varepsilon_{t+1} = \varepsilon_t + \gamma (\varepsilon - \text{err}_t^{\varepsilon})
\end{equation*}
where 
\begin{equation*}
    \text{err}_t^{\varepsilon} = \begin{cases}
    1 & \text{if $y_t\notin\Gamma_t^{\varepsilon_t}$}\\
    0 & \text{otherwise}.
    \end{cases}
\end{equation*}

The theoretical guarantees of ACI are somewhat different from CP, which guarantees that errors are independent and happen with at most a user specified probability $\varepsilon$. What ACI promises is weaker, but still useful, in that it provides finite sample guarantees on the coverage frequency: For any positive integer $T$, and learning rate $\gamma>0$, we have that
\begin{equation}
    \label{eq:aciFiniteSampleBound}
    \begin{matrix}
        \bigg|\frac{1}{T}\sum_{t=1}^T\text{err}_t-\varepsilon\bigg|\leq\frac{\max\{\varepsilon_1, 1-\varepsilon_1\} + \gamma}{\gamma T} & (a.s.)
    \end{matrix}
\end{equation}
In words, the absolute deviation from the desired error-rate, for a finite sample size $T$ will almost surely not exceed $\frac{\varepsilon_1+\gamma}{\gamma T}$ (assuming $\epsilon_1 \leq1/2$), which in particular converges to 0 as $T \to\infty$. Thus, ACI ensures that a conformal predictor $\Gamma$ is asymptotically valid, even if the data is not exchangeable.

In time-series forecasting, we are often interested in predicting more than just one step ahead, which is known as multi-step ahead forecasting. In this situation, we can not use ACI directly because the online update requires knowledge of the errors immediately, but if we predict $h$ steps ahead, we only learn whether our prediction was correct after $h$ time steps have passed. 

The main contribution of this paper is to adapt the ACI algorithm to multi-step ahead forecasts in the online setting. To the best of my knowledge, this has not been done before. ACI has been suggested as part of the adaptive ensemble batch multi-input multi-output conformalised quantile regression (AEnbMIMOCQR) in \cite{sousa2022general} which uses bootstrapping, similar to the ensemble batch prediction intervals (EnbPI) algorithm introduced by \cite{xu2021conformal}. Another approach to conformal multi-step ahead forecasting, taken in \cite{schlembach2022conformal}, uses the weighted quantiles method of \cite{barber2023conformal} together with an inductive conformal predictor in the offline mode. Empirical validation supports the method, but no theoretical guarantees are presented. 
The CopulaCPTS method of \cite{sun2022copula} comes with finite sample coverage guarantees, but their focus is on the setting where data consists of many independent time series, which is different from our aim.

Common for these works is that they apply some form of inductive conformal predictor, which requires data splitting. 
Our approach differs by being entirely online, avoiding data splitting, and can thus be used together with any online conformal predictor that produces multi-step ahead forecasts. As we shall see, our proposed method is able to use different target miss-coverage levels for different prediction steps, allowing a user to balance between high coverage and tight prediction sets when predicting far ahead, which to the best of my knowledge has not been a feature of any previous conformal method.

To illustrate the method in practice, we provide numerical examples in \sectionref{sec:example}. The examples provided are not intended as experimental results, but merely serve to illustrate the proposed method in practice. Almost any machine learning method can be made into a conformal predictor, and any online conformal predictor is compatible with the proposed multi-step ahead ACI. In our examples, we use an adapted version of the conformalised ridge regression (CRR) algorithm (see \algorithmref{alg:mimo-crr}) to produce multi-step ahead predictions online in a computationally efficient manner. We make no claim that this adaptation is particularly well suited for multi-step ahead time-series forecasting, but it serves to illustrate the multi-step ahead ACI procedure.

The rest of the paper is organised as follows. We briefly discuss different strategies for multi-step ahead forecasting, and introduce a modified version of the confromalised ridge regression algorithm in \sectionref{sec:MIMO}. Our main contribution, the multi-step ahead ACI algorithm, is introduced in \sectionref{sec:MSA-ACI}, where we also discuss its finite sample coverage guarantees and asymptotic properties, both at individual forecast steps and overall. The numerical examples, using a publicly available electricity demand dataset, in \sectionref{sec:example} serve to illustrate the method in practice. We show the effect of using different target miss-coverage rates and/or learning rates at different time-steps, which could be useful in practice. \sectionref{sec:conclusions} concludes.

\section{Multi-step ahead time-series forecasting}\label{sec:MIMO}

In a multi-step ahead forecasting situation, we must take care to choose an effective forecast strategy. Suppose we have a time series $(w_1,w_2,\dots,w_n)$, and want to estimate a function $\mathbb{R}^p\to\mathbb{R}^h$ to predict the next $h$ values based on $p$ lagged values. We could include exogenous variables as well, which increases the dimension of the domain, but for simplicity, let us ignore this possibility here. 
The multi-input multi-output (MIMO) strategy converts the forecasting problem to a supervised learning problem with multiple targets:
\begin{equation*}
    \left(
        \begin{array}{@{}cccc|ccc@{}}
            w_1 & w_2 & \dots & w_p & w_{p+1} & \dots & w_{p+h} \\
            w_2 & w_3  &\dots & w_{p+1} & w_{p+2} & \dots & w_{p+h+1} \\
            \vdots & \vdots & \ddots & \vdots & \vdots & \ddots & \vdots \\
            w_{n-h-p} & w_{n-h-p+1} & \dots & w_{n-h-1} & w_{n-h} & \dots & w_n
        \end{array}
    \right)
\end{equation*}
with the lagged values on the left-hand side of the matrix, and the targets on the right-hand side. We will refer to the $t$th row at the left-hand side of the matrix as the object $x_t$, and  the $t$th row at the right-hand side as the label $y_t$. Empirical studies \citep{taieb2012review, xiong2013beyond} have shown that MIMO strategies tend to outperform recursive strategies that use one-step ahead forecasts to produce the next forecast, which tends to accumulate prediction errors. This is however not an absolute rule, and there are situations where recursive strategies may be a better option. 

The choice of strategy will thus depend on the particular problem at hand. For the numerical examples in \sectionref{sec:example}, we have chosen to use the MIMO strategy. In particular, a MIMO version of ridge regression to be defined below.

\subsection{MIMO conformalised ridge regression}\label{sec:MIMO-CRR}

We adapt the conformalised ridge regression (CRR) algorithm (Algorithm 2.4 \citealp{alrw2}) to handle multi-step ahead predictions, resulting in the MIMO-CRR algorithm, which is described in \algorithmref{alg:mimo-crr}. The adaptation is simple, but I was unable to find any online MIMO conformal predictor in the literature, so it seems useful to present the algorithm in full. Recall that $(a_{ij})$ is a convenient shorthand notation for the matrix whose elements are $a_{ij}$ and that $I_p$ is the $p\times p$ identity matrix.

\begin{algorithm2e}[H]
\caption{MIMO-CRR}\label{alg:mimo-crr}
\KwData{Ridge parameter $a\geq0$, vector of significance levels $\boldsymbol{\varepsilon} = (\varepsilon_i)_{i=1}^h\in(0,1)^h$, training set $(x_i,y_i)\in\mathbb{R}^p\times\mathbb{R}^h, i=1,\dots n-1$ and a test object $x_n\in\mathbb{R^p}$.}
Set $X_n:=(x_1,\dots,x_n)^T$\;
set $H_n := X_n(X_n^TX_n+aI_p)^{-1}X_n^T$\;
set $C := I_n-H_n$\;
set $A := (a_{ij}):=C(y_1,\dots,y_{n-1},0)^T$ and $B := (b_{ij}):=C(0,\dots,0,1)^T$, with $0,1$ understood as vectors of length $h$ whose elements are all zeros and ones respectively\;
\For{{$i=1,\dots,h$}}{
    \For{$j=1,\dots,n-1$}{
        \eIf{$b_{ni} > b_{ji}$}{
            set $u_j:=l_j:=(a_{ji}-a_{ni})/(b_{ni}-b_{ji})$
        }{
        set $l_j=-\infty$ and $u_j=\infty$
        }
    }
    sort $u_1,\dots,u_{n-1}$ in the ascending order obtaining $u_{(1)}\leq\dots\leq u_{(n-1)}$\;
    sort $l_1,\dots,l_{n-1}$ in the ascending order obtaining $l_{(1)}\leq\dots\leq l_{(n-1)}$\;
    set $\Gamma_i^{(\varepsilon_i)} := [l_{(\lfloor (\varepsilon_i/2)n \rfloor)}, u_{(\lceil (1-\varepsilon_i/2)n \rceil)}]$\;
}
output $(\Gamma_1^{(\varepsilon_1)}, \dots, \Gamma_h^{(\varepsilon_h)})$ as prediction set. 
\end{algorithm2e}

Note that $l_{(\lfloor (\varepsilon_i/2)n \rfloor)}, u_{(\lceil (1-\varepsilon_i/2)n \rceil)}$ is defined only when $\varepsilon_i \geq 2/n$ for all $i$. The MIMO-CRR can use lagged values together with exogenous variables to produce prediction sets online. It should be remarked that the output set is a $h$-tuple whose $i$th element is a prediction interval for $y_{n+i-1}$. It is also worth noting that the matrix $(X_n^TX_n+aI_p)^{-1}$ can be updated efficiently online as new examples arrive by the Sherman–Morrison formula, see e.g. \citep{bartlett1951inverse}, to avoid the costly matrix inversion at each step.

\section{Multi-step ahead adaptive conformal inference}\label{sec:MSA-ACI}

The ACI algorithm was introduced in \cite{gibbs2021adaptive}. It works in the online setting by a simple update of the significance level. Here we adapt it to handle multi-step ahead forecasts, allowing for different confidence levels and learning rate at different prediction steps.

First, because there is in general a trade-off to be made between high confidence and tight prediction intervals, it may be advantageous to allow for forecasts that lie further in the future to have lower confidence. Of course, we would like to be very confident, but if the prediction intervals are too wide, they may not be useful. Similarly, we may wish to set the learning rate of ACI at different levels depending on the forecast step. There is a trade-off between stability and adaptability when choosing learning rate, with large learning rates enabling more adaptability at the cost of decreased stability. Gibbs and Cand\`es suggest that in environments with large distribution shifts, the learning rate should be chosen higher, which makes intuitive sense. In a multi-step ahead setting, it would therefore seem natural to set the learning rates successively larger the further ahead we predict.
Thus, a vector of desired miss-coverage rates
\begin{equation}
    \label{eq:epsilon}
    \boldsymbol{\varepsilon} = (\varepsilon_1, \dots, \varepsilon_h)
\end{equation}
is specified by the user, together with a vector of learning rates
\begin{equation}
    \label{eq:gamma}
    \boldsymbol{\gamma} = (\gamma_1, \dots, \gamma_h).
\end{equation}

From now on, a bold symbol denotes a vector. At time $t$, after we have made our prediction, and the true label is revealed, we have access to the values $y_{t-h+1,1},\dots,y_{t,1}$, we can then fully observe the correctness of our predictions made at time $t-h+1$, but also partially the correctness of the predictions made since. Denote by $l_{t,i}$ and $u_{t,i}$ the lower and upper bounds of our prediction intervals, with $i=1,\dots,h$. Consider the matrices,
\begin{equation}
    \label{eq:Ut}
    U_t = 
    \begin{pmatrix}
    u_{t,1} & u_{t,2} & \dots & u_{t,h} \\
    u_{t-1,1} & u_{t-1,2} & \dots & u_{t-1,h} \\
    \vdots & \vdots & \ddots & \vdots \\
    u_{t-h+1,1} & u_{t-h+1,2} & \dots & u_{t-h+1,h} \\
    \end{pmatrix}
\end{equation}
and
\begin{equation}
    \label{eq:Lt}
    L_t = 
    \begin{pmatrix}
    l_{t,1} & l_{t,2} & \dots & l_{t,h} \\
    l_{t-1,1} & l_{t-1,2} & \dots & l_{t-1,h} \\
    \vdots & \vdots & \ddots & \vdots \\
    l_{t-h+1,1} & l_{t-h+1,2} & \dots & l_{t-h+1,h} \\
    \end{pmatrix}.
\end{equation}
The diagonal elements are the lower and upper bounds of the predictions made at time $t-h+1, \dots t$ for the value $y_t$. Thus, the vector of errors
\begin{equation}
    \label{eq:err}
    \textbf{err}_t = (\text{err}_{t,1}, \text{err}_{t-1,2},  \dots \text{err}_{t-h+1,h})
\end{equation}
can be evaluated as
\begin{equation*}
    \neg\bigg(\text{diag}(L_t) \leq y_{t,1} \leq \text{diag}(U_t)\bigg)
\end{equation*}
where “$\neg$” denotes logical negation, and the comparisons are made element-wise.
We can then update our control input vector 
\begin{equation}
    \label{eq:controlInput}
    \boldsymbol{\varepsilon}_t = (\varepsilon_{t,1}, \dots, \varepsilon_{t,h})
\end{equation}
by
\begin{equation}
    \label{eq:aciUpdate}
    \boldsymbol{\varepsilon}_{t+1} = \boldsymbol{\varepsilon}_t + \boldsymbol{\gamma} (\boldsymbol{\varepsilon} - \textbf{err}_t)
\end{equation}
with the vector multiplication understood as element-wise multiplication. Initially, we have an issue. For $t=1,\dots,h$, we do not yet know the entire error vector. To get around this, we initialise the error vector as $\textbf{err}_1 = \boldsymbol{\varepsilon}$, which is a fixed point for the ACI update \equationref{eq:aciUpdate}, so that the significance level at hour $i$ is kept fixed until the first relevant error arrives. When errors arrive, they are added in their respective places until we have observed an entire error vector, which can then be used in \equationref{eq:aciUpdate} without any modification. This means that e.g. $\varepsilon_{t,h}$ is kept fixed for $t=1,\dots,h$, and is only then adaptively updated.

Because the multi-step ahead forecasting protocol here described is simply applying ACI independently at each prediction step, using the incoming errors to update as they arrive, it inherits all theoretical guarantees of ACI. In particular, for each forecast step, assuming $\gamma_i>0$:
\begin{equation}
    \label{eq:mhaciFiniteSampleBound}
    \begin{matrix}
        \bigg|\frac{1}{T}\sum_{t=i}^T\text{err}_{t,i}-\varepsilon_i\bigg| \leq \frac{\max\{\varepsilon_{i,i}, 1-\varepsilon_{i,i}\} + \gamma_i}{\gamma_i T}, & i=1,\dots,h & (a.s.).
    \end{matrix}
\end{equation}
Note that the summation begins at $t=i$, which reflects that $\varepsilon_{t,i}$ is kept fixed until we observe the relevant error.
The overall error rate is $\frac{1}{hT}\sum_{t=1}^T\sum_{i=1}^h\text{err}_{t,i}$, and combining the above inequalities, yields the finite sample bound

\begin{equation}
    \label{eq:mhaciFiniteSampleBoundOverall}
    \begin{aligned}
        \bigg|\frac{1}{T}\sum_{t=h}^T(\frac{1}{h}\sum_{i=1}^h\text{err}_{t,i}) - \frac{1}{h}\sum_{i=1}^h\varepsilon_i\bigg|
        &= \frac{1}{h}\bigg|\sum_{i=1}^h(\frac{1}{T}\sum_{t=h}^T\text{err}_{t,i}) - \sum_{i=1}^h\varepsilon_i\bigg|\\
        &= \frac{1}{h}\bigg|\sum_{i=1}^h(\frac{1}{T}\sum_{t=h}^T\text{err}_{t,i} - \varepsilon_i)\bigg|\\
        &\leq \frac{1}{h}\sum_{i=1}^h\bigg|\frac{1}{T}\sum_{t=h}^T\text{err}_{t,i} - \varepsilon_i\bigg|\\
        &\leq \frac{1}{h}\sum_{i=1}^h \frac{\max\{\varepsilon_{h,i}, 1-\varepsilon_{h,i}\} + \gamma_i}{\gamma_i T} ~(a.s.)
    \end{aligned}
\end{equation}
for the overall error rate beyond time $h$, assuming $\boldsymbol{\gamma} >0$ (elementwise). The second last line follows from the triangle inequality, and the last line follows from \equationref{eq:mhaciFiniteSampleBound}. Note that the summation begins at $t=h$, since before this point, $\varepsilon_{t,h}$ is kept fixed, so we can not apply \equationref{eq:mhaciFiniteSampleBound}. 

In particular, both the inequalities in \equationref{eq:mhaciFiniteSampleBound} and that in \equationref{eq:mhaciFiniteSampleBoundOverall} imply that the error rate at each prediction step as well as the overall error rate converges to the desired ones as $T\to\infty$. To put it in the terminology of \cite{alrw2}, the resulting confidence predictor is asymptotically valid at each prediction step (considered as a confidence predictor of just that step) as well as overall.

Note that nothing prevents $\varepsilon_{t,i} < 0$ which would result in an infinite prediction interval for $y_{t,i}$. This can however be avoided by enforcing a lower bound for each control input. If we use the MIMO-CRR algorithm, we have to enforce $\varepsilon_{t,i} \geq 2/n$ to ensure that the prediction sets are defined. This also ensures that the prediction sets are finite unless $b_{ni} > b_{ji}$. It is important to note that the finite-sample bounds \equationref{eq:mhaciFiniteSampleBound} and \equationref{eq:mhaciFiniteSampleBoundOverall} may be violated if we do this, but in practice it may be prudent to avoid infinite prediction intervals. It may however depend on the use case. 

Note that the multi-step ahead ACI algorithm is agnostic to the forecast strategy employed. Our numerical illustrations use the MIMO strategy, but it would work equally well to employ the recursive strategy described in \sectionref{sec:MIMO}.

\section{Example: electricity demand forecast}\label{sec:example}

We illustrate the multi-step ahead ACI wrapped around MIMO-CRR on the Victoria electricity demand dataset, which is available through the MAPIE GitHub repository\footnote{https://raw.githubusercontent.com/scikit-learn-contrib/MAPIE/master/\\examples/data/demand\_temperature.csv} \citep{taquet2022mapie}. The dataset consists of hourly electricity demands together with the temperature at each time step. We enrich the dataset with date information (week of year, weekday, hour of day). Each object $x_t$ consists of this information and 24 lagged values of the demand. The forecast horizon $h$ is set to 5, so each label $y_t = (y_{t,1}, \dots, y_{t,5})$, where $y_{t,i}$ is the demand at time $t+i-1$. We use 477 historical values as initial training set, and tune the ridge parameter with generalised cross-validation \citep{golub1979generalized}. We set $\boldsymbol{\varepsilon_1} = \boldsymbol{\varepsilon}$ in all examples, that is, our first vector of significance levels, is set to be equal to the vector of target error rates.

We present three cases. First, the target error and learning rates are kept the same over the entire forecast horizon. Second, the target error rate is different for different forecast hours, allowing for a more casual prediction when we predict far ahead, which should keep the prediction sets narrower. In our last example, we also set the learning rate higher for further ahead forecasts; the intuition being that greater adaptability may be required when we predict far ahead. 

It should be noted that these examples are not meant as experimental results, but merely serve to illustrate the proposed method in action. As noted in the introduction, there are other algorithms for conformal multi-step ahead forecasting, some of which could be adapted to the online setting to facilitate comparison, but this is left as future work for now.

\subsection{Same target coverage and learning rate for all}

\begin{figure}[h]
\floatconts
{fig:example1}
{\caption{Example 1. Same target coverage rate and learning rate for all predictions.}}
{%
\subfigure[Prediction sets for hour 1 and 5.]{%
\label{fig:same_eps_same_gam_plot}
\includegraphics[width=0.8\linewidth]{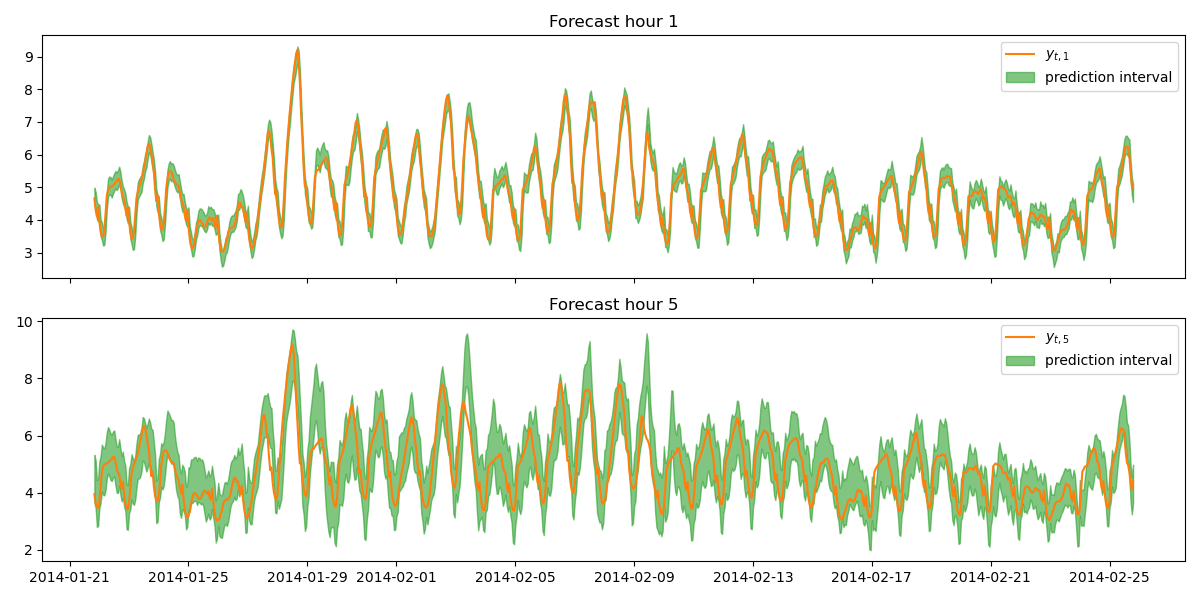}
}\qquad 
\subfigure[Control inputs for ACI and the interval width at each hour.]{%
\label{fig:same_eps_same_gam_control}
\includegraphics[width=0.8\linewidth]{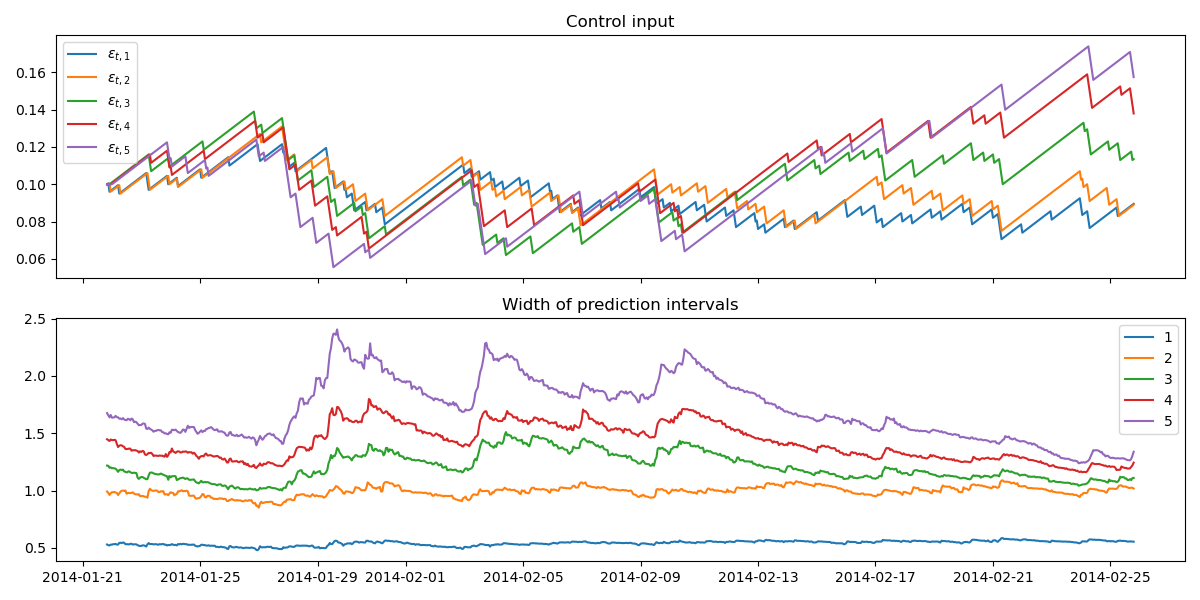}
}
}
\end{figure}

Our first example illustrates the potential issue with using the same confidence level at all prediction steps. We set $\varepsilon_i = 0.1$ and $\gamma_i = 0.005$ for $i=1,\dots,5$. \figureref{fig:same_eps_same_gam_plot} shows the prediction intervals for hour 1 and hour 5. The average width of the intervals together with the empirical miss-coverage rate for each hour are presented in \tableref{tab:same_eps_same_gam}. The control inputs $\varepsilon_{t,i}$ for $i=1,\dots,5$ together with the width of the prediction intervals for each hour are shown in \figureref{fig:same_eps_same_gam_control}. Overall, we see that the further ahead predictions are rather conservative, and the prediction sets tend to be quite large.

\clearpage

\subsection{Different target coverage but same learning rate}

\begin{figure}[h]
\floatconts
{fig:example2}
{\caption{Example 2. Different target error rates for different forecast hours, but same learning rate. Note the difference in scale on the $y$-axis compared to Figure \ref{fig:same_eps_same_gam_control}.}}
{%
\subfigure[Prediction sets for hour 1 and 5.]{%
\label{fig:diff_eps_same_gam_plot}
\includegraphics[width=0.8\linewidth]{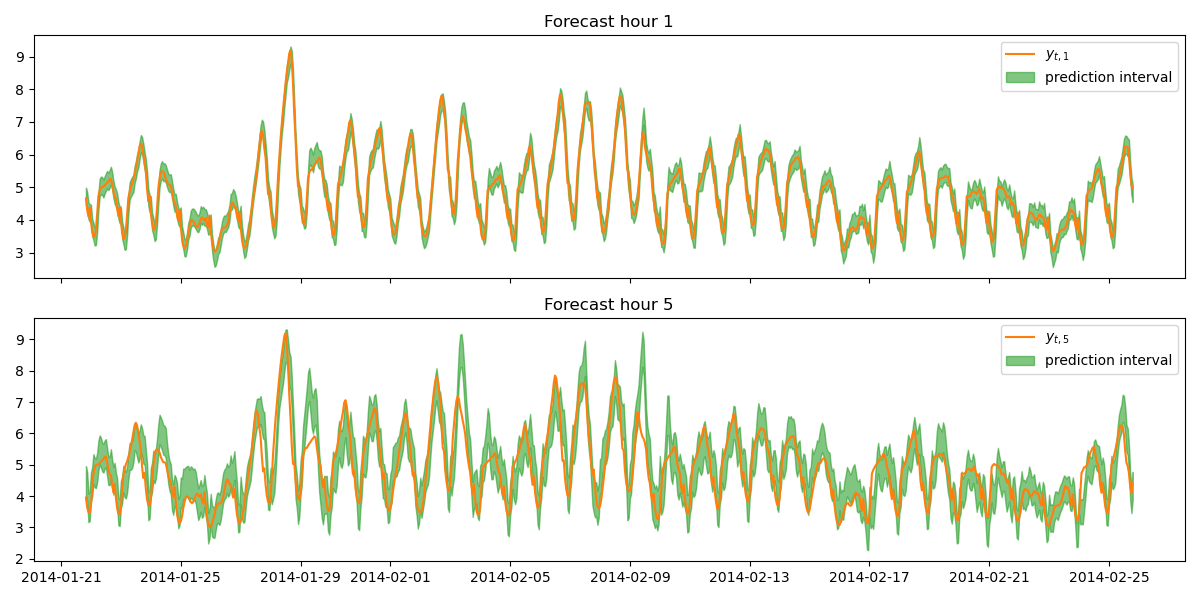}
}\qquad 
\subfigure[Control inputs for ACI and the interval width at each hour.]{%
\label{fig:diff_eps_same_gam_control}
\includegraphics[width=0.8\linewidth]{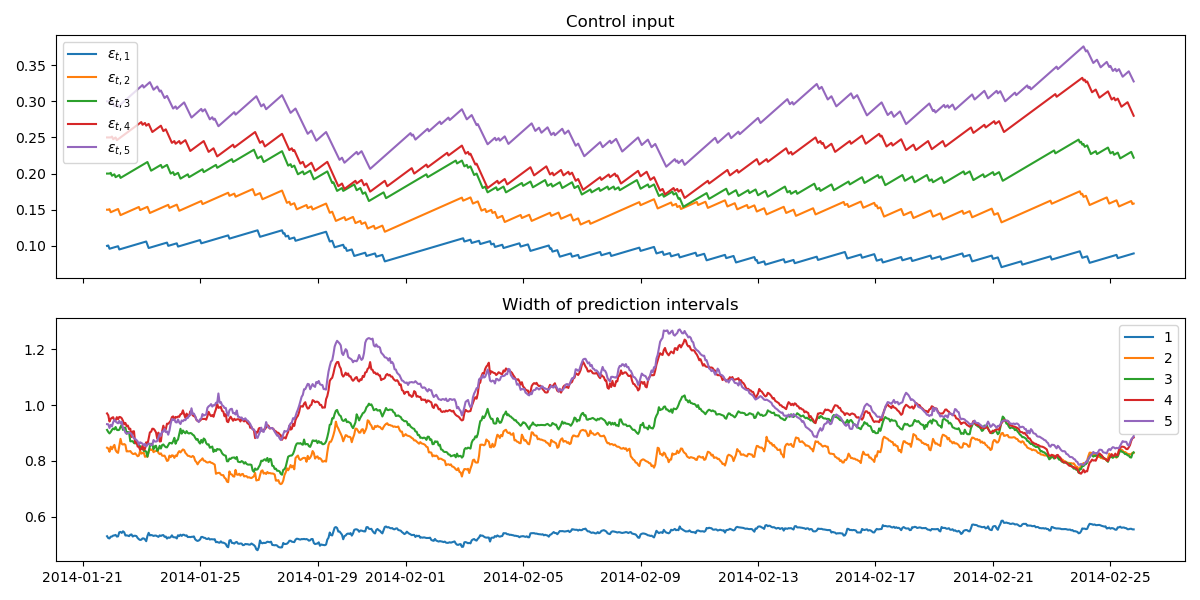}
}
}
\end{figure}

Our second example illustrates the effect of setting the target coverage rate to different values while keeping the learning rates fixed. We keep $\gamma_i = 0.005$ but set $\boldsymbol{\varepsilon} = (0.1, 0.15, 0.2, 0.25, 0.3)$. Figures \ref{fig:diff_eps_same_gam_plot} and \ref{fig:diff_eps_same_gam_control} together with \tableref{tab:diff_eps_same_gam} summarise the result. The target miss-coverage rate is approximately achieved. Because we decrease the confidence level for further ahead predictions, the prediction sets are much narrower than when we kept the confidence level fixed.

\clearpage

\subsection{Different target coverage and learning rates}

\begin{figure}[h]
\floatconts
{fig:example3}
{\caption{Example 3. Different target coverage and learning rates for different forecast hours. Note the difference in scale on the $y$-axis compared to \figureref{fig:same_eps_same_gam_control}.}}
{%
\subfigure[Prediction sets for hour 1 and 5.]{%
\label{fig:diff_eps_diff_gam_plot}
\includegraphics[width=0.8\linewidth]{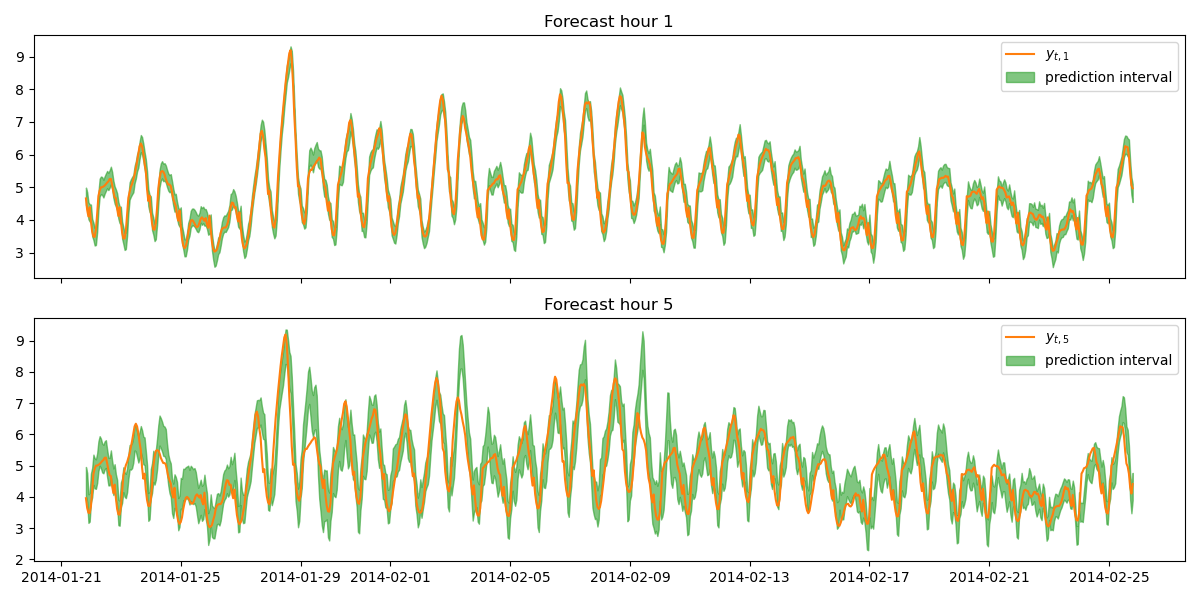}
}\qquad 
\subfigure[Control inputs for ACI and the interval width at each hour.]{%
\label{fig:diff_eps_diff_gam_control}
\includegraphics[width=0.8\linewidth]{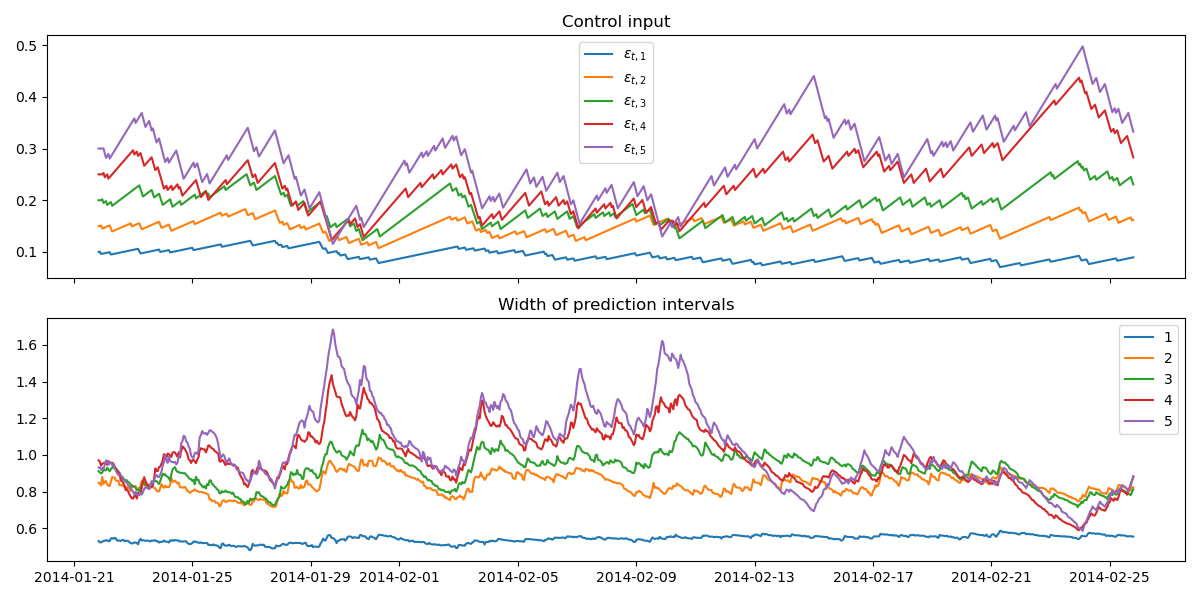}
}
}
\end{figure}

Our last example uses different target coverage rates and different learning rates for different forecasts. Specifically we have
\begin{equation*}
    \begin{aligned}
        \boldsymbol{\varepsilon} &= (0.1, 0.15, 0.2, 0.25, 0.3) \\
        \boldsymbol{\gamma} &= (0.005, 0.007, 0.009, 0.011, 0.013).
    \end{aligned}
\end{equation*}
Figures \ref{fig:diff_eps_diff_gam_plot}-\ref{fig:diff_eps_diff_gam_control} together with \tableref{tab:diff_eps_diff_gam} shows that the target error rate is achieved more closely by varying the learning rate over the different prediction hours. We see from \figureref{fig:diff_eps_diff_gam_control} that the control inputs vary dramatically for hour 4 and 5. If the learning rates are chosen too high, there is a risk of loosing stability.

\clearpage

\begin{table}[h]
    \centering
    \begin{tabular}{c|cccccc}
        hour & 1 & 2 & 3 & 4 & 5 & overall \\
         \hline
        $\varepsilon$ & 0.1 & 0.1 & 0.1 & 0.1 & 0.1 & 0.1 \\
        error rate & 0.102 & 0.102 & 0.0964 & 0.0905 & 0.0869 & 0.0957 \\
        avg length & 0.541 & 0.994 & 1.21 & 1.49 & 1.71 & 1.17
    \end{tabular}
    \caption{Example 1. Same target coverage rate and learning rate for all predictions. The predictions at hour 5 are conservative.}
    \label{tab:same_eps_same_gam}
\end{table}

\begin{table}[h]
    \centering
    \begin{tabular}{c|cccccc}
        hour & 1 & 2 & 3 & 4 & 5 & overall \\
         \hline
        $\varepsilon$ & 0.1 & 0.15 & 0.2 & 0.25 & 0.3 & 0.2 \\
        error rate & 0.102 & 0.148 & 0.194 & 0.243 & 0.295 & 0.196 \\
        avg length & 0.541 & 0.837 & 0.905 & 0.997 & 1.01 & 0.858
    \end{tabular}
    \caption{Example 2. Different target error rates for different forecast hours, but same learning rate. Control inputs for ACI and the interval width at each hour.}
    \label{tab:diff_eps_same_gam}
\end{table}

\begin{table}[h]
    \centering
    \begin{tabular}{c|cccccc}
        hour & 1 & 2 & 3 & 4 & 5 & overall \\
         \hline
        $\varepsilon$ & 0.1 & 0.15 & 0.2 & 0.25 & 0.3 & 0.2 \\
        error rate & 0.102 & 0.148 & 0.195 & 0.246 & 0.298 & 0.198 \\
        avg length & 0.541 & 0.841 & 0.919 & 0.980 & 1.03 & 0.861
    \end{tabular}
    \caption{Example 3. Different target coverage and learning rates for different forecast hours.}
    \label{tab:diff_eps_diff_gam}
\end{table}

\section{Conclusions}\label{sec:conclusions}

We have adapted the CRR algorithm to deal with multi-step ahead time-series forecasts using the MIMO strategy, and used an adapted version of the ACI algorithm to vary the significance levels at each prediction step to approximately achieve the desired error rate. Our adaptation allows for different target error rates, and different learning rates for different prediction steps. The precise tuning of suitable target error and learning rates will be use case specific and may depend on potential periodicity and/or trends in the time-series at hand.

The Conformal PID control method, introduced in \cite{angelopoulos2024conformal}, may be viewed as a generalisation of ACI, where ACI is the special case of using only the P part of their method. Future work could investigate the utility of adapting conformal PID to the multi-step ahead forecast setting. More generally, any method that uses the significance level as control input in the online setting, to achieve a target error rate could be put in our framework.

While the MIMO-CRR algorithm can handle multi-step ahead forecasts, we make no claims that it is a particularly good choice for electricity demand forecasting, but is merely used to illustrate the multi-step ahead ACI procedure. The choice of time-series forecasting algorithm will always depend on the problem at hand. Our purpose here has not been to test the MIMI-CRR algorithm per se. As an online conformal regression algorithm, it is computationally efficient, at least for modest object dimensions $p$. In cases when $p$ is large, the matrix inversion step may be computationally intractable, say if we have a very large number of lags, and/or many exogenous variables. In such cases, it may be desirable to use kernel ridge regression instead, which can be conformalised similarly. Future work may consider the use of the resulting conformalised kernel ridge regression in MIMO multi-step ahead forecasting. Any online CP method may be used together with multi-step ahead ACI to achieve approximately the desired coverage frequency. As always, the efficiency will depend on the underlying machine learning method.

\acks{I should like to thank the anonumous reviewers for their valuable suggestions for improvements. I also acknowledge the Swedish Knowledge Foundation and industrial partners for financially supporting the research and education environment on Knowledge Intensive Product Realisation SPARK at Jönköping University, Sweden. Project: PREMACOP grant no. 20220187.}


\newpage

\vskip 0.2in
\bibliography{sample}

\end{document}